\documentclass[final]{cvpr}

\usepackage{times}
\usepackage{epsfig}
\usepackage{graphicx}
\usepackage{amsmath}
\usepackage{amssymb}

\usepackage{bbding}
\usepackage{pifont}
\usepackage{wasysym}
\usepackage{amssymb}
\usepackage{multirow}
\usepackage[pagebackref=true,breaklinks=true,colorlinks,bookmarks=false]{hyperref}

\usepackage[margin=1in]{geometry}
\usepackage[utf8]{inputenc}
\PassOptionsToPackage{hyphens}{url}
\usepackage{cleveref}
\makeatletter
\@namedef{ver@everyshi.sty}{}
\makeatother
\usepackage{tikz}
\usepackage{tikz,tikz-qtree}
\usepackage[numbers]{natbib}

\def\cvprPaperID{3327} % *** Enter the CVPR Paper ID here
\def\confYear{CVPR 2021}
\newcommand{\framework}{\textit{F-IF Framework}\xspace}

\begin{document}
	
	%%%%%%%%% TITLE
	\title{On the Limitations of Denoising Strategies as Adversarial Defenses}
	
	\author{Zhonghan Niu\\
		 Nanjing University\\
		{\tt\small niuzh@smail.nju.edu.cn}
		\and
		Zhaoxi Chen\\
		Tsinghua University\\
		{\tt\small frozen.burning@gmail.com}
		\and
		Linyi Li\\
		UIUC\\
		{\tt\small linyi2@illinois.edu}
		\and
		Yubin Yang\\
		Nanjing University\\
		{\tt\small yangyubin@nju.edu.cn}
		\and
		Bo Li\\
		UIUC\\
		{\tt\small lbo@illinois.edu}
		\and
		Jinfeng Yi\\
		JD AI Research\\
		{\tt\small yijinfeng@jd.com}
	}
	
	\maketitle

	%%%%%%%%% ABSTRACT
    \begin{abstract}
    As adversarial attacks against machine learning models have raised increasing concerns, many denoising-based defense approaches have been proposed. In this paper, we summarize and analyze the defense strategies in the form of symmetric transformation via data denoising and reconstruction (denoted as $F+$ inverse $F$, or $\framework$). In particular, we categorize these denoising strategies from three aspects (\textit{i.e.} denoising in the spatial domain, frequency domain, and latent space, respectively). Typically, defense is performed on the entire adversarial example, both image and perturbation are modified, making it difficult to tell how it defends against the perturbations. To evaluate the robustness of these denoising strategies intuitively, we directly apply them to defend against adversarial noise itself (assuming we have obtained all of it), which saving us from sacrificing benign accuracy. Surprisingly, our experimental results show that even if most of the perturbations in each dimension is eliminated, it is still difficult to obtain satisfactory robustness. Based on the above findings and analyses, we propose the adaptive compression strategy for different frequency bands in the feature domain to improve the robustness. Our experiment results show that the adaptive compression strategies enable the model to better suppress adversarial perturbations, and improve robustness compared with existing denoising strategies.
    \end{abstract}
	
%%%%%%%%% BODY TEXT
    \section{Introduction}
    % As the use of machine learning models increases, security has become a critical property of the deployed intelligent system. However,
    Deep neural networks (DNNs) have been shown to be vulnerable to adversarial perturbations\cite{FGSM,szegedy2014intriguing}. Recent works \cite{brendel2018decisionbased,papernot2017practical,liu2017delving} successfully attack deep models with human-imperceptible adversarial noise. In particular, malicious adversarial noise is injected into the inputs to mislead the machine learning models with high confidence. Since these deep models have been widely used in safety-critical scenarios such as face recognition, autonomous driving, and medical care, it is urgent to improve the robustness of neural networks, and thus the defense techniques have become crucial for secure deep learning. In recent years, lots of efforts have been devoted to improving the robustness against adversarial attacks. Since the adversarial examples are constructed by directly adding well-craft adversarial noise to the legitimate images. Therefore, an intuitive defense idea is to adopt denoising strategies, preprocess these adversarial examples to eliminate as much adversarial noise as possible before feeding them into the model. 
    \begin{figure*}[h]
	\begin{center}
		\includegraphics[width=0.7\linewidth]{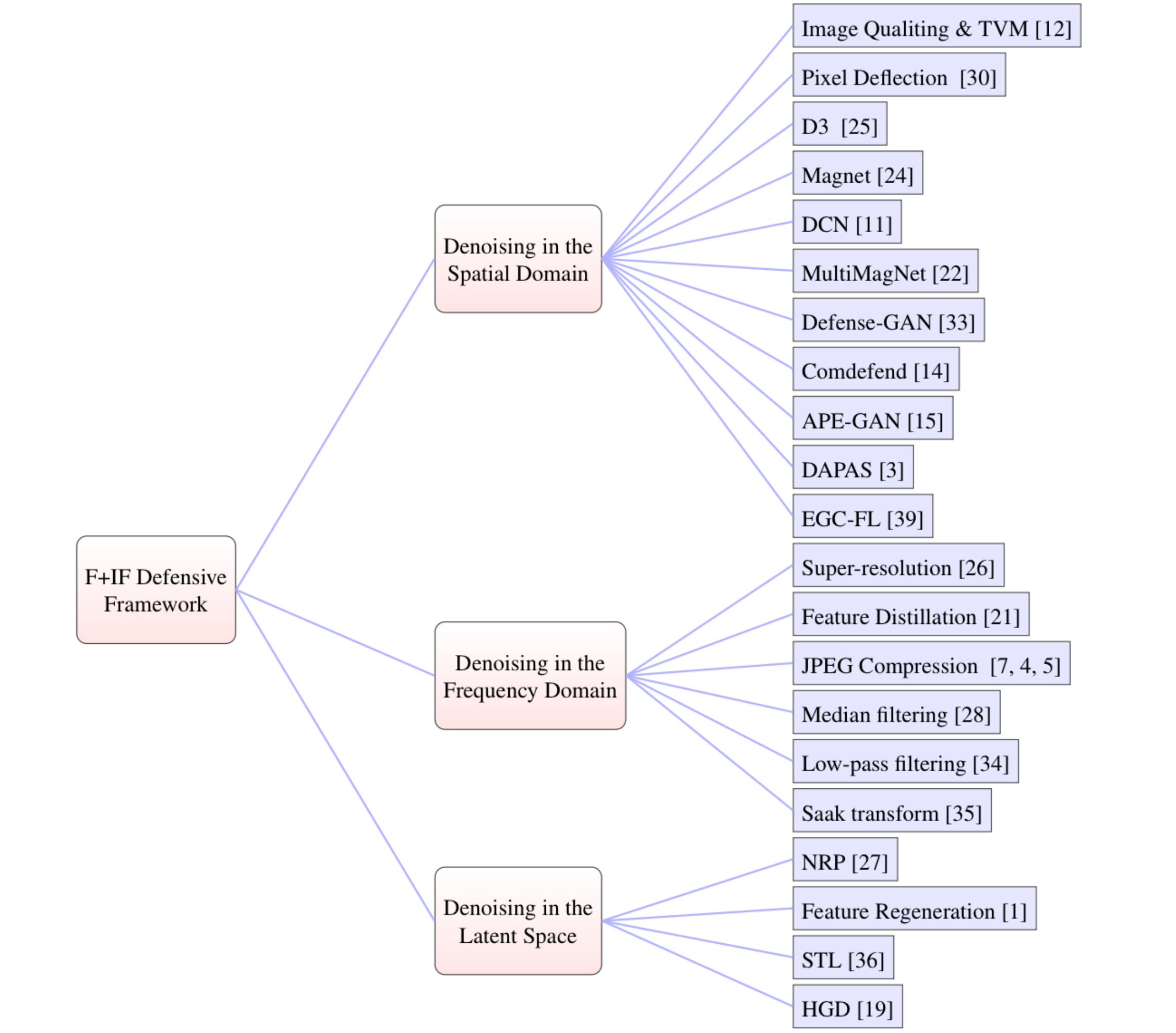}
	\end{center}
	% 	\vspace{-1em}
	\caption{Summarization of the denoising-based defense methods via the \framework.}
	\label{fig:Tree}
\end{figure*}
    In this paper, we leverage the \framework (a symmetric transformation framework via image denoising and reconstruction), summarize the existing denoising-based defense methods, and shed new light on improving the defense robustness. First, from a novel perspective, we categorized the denoising strategies from three main aspects: denoising in the spatial domain, frequency domain, and latent space. Second, based on the strong assumption that the full adversarial noise $\eta$ (see Equ~\ref{eq:perturbation}) are obtainable, these defense strategies can be directly applied to the perturbation itself. Without sacrificing the benign accuracy, we can evaluate the robustness of each strategy quantitatively. Surprisingly, the results in Sec.~\ref{sub:spatial_analysis}, \ref{sub:frequency_analysis}, \ref{sub:latent_analysis} show that the denoising strategies are not as effective as expected. Meanwhile, we further analyze the characteristics of the perturbations to acquire a better understanding. Finally, based on the above results and analyses, we suggest adopting adaptive compression strategies for various frequency bands in the feature domain to improve the robustness of machine learning models. Our results demonstrate the necessity of the adaptive compression strategy as it empowers models to suppress adversarial perturbations and achieve higher robustness than the existing denoising strategies we have investigated. The major contributions of this paper can be summarized as follows:

% In this work, we leveraged the \framework, a symmetric transformation framework via image denoising and reconstruction, to categorize these denoising strategies from a novel perspective (see Fig.~\ref{fig:Tree}). Within the \framework, defenses that deploy denoising strategies fall into one of three main categories: denoising in the spatial domain~\cite{qualiting,D3,pixel-deflection,DCN,magnet,MultiMagNet,DAPAS,Comdefend,APE-GAN,Defense-GAN,EGC-FL}, frequency domain~\cite{SR,Feature-Distillation,JPEG1-study,JPEG2-keep,JPEG3-Shield,Medianfilter,Low-pass,Saak-Transform}, and latent space~\cite{NRP,Feature-Regeneration,STL,HGD}.
% Second, based on the assumption that the full perturbation (see Eq.~\ref{eq:perturbation}) of the adversarial attack is obtainable, we extensively evaluate these defenses in a consistent manner by eliminating a certain portion of adversarial noise against the attack powered by FGSM~\cite{FGSM} or PGD~\cite{PGD}, which enables us to evaluate the defense efficiency without sacrificing benign accuracy.

    \begin{itemize}
	\item We summarize the denoising-based defense strategies via \framework, and analyze them from three aspects: denoising in the spatial domain, frequency domain, and latent space, respectively.
	
	\item To evaluate the robustness of these denoising strategies more intuitively, we directly apply these strategies to defend against adversarial noise. Surprisingly, all the results indicate the limitations of denoising strategies.
	
	\item Our observations and analyses shed new light on future defense research, then we propose a promising adaptive compression defense strategy. The \textit{Adaptive Compression and reconstruction Model} (ACM) construct based on this strategy achieves a much higher robustness performance than existing denoising-based methods.
    \end{itemize}
	
	\section{Summarization and Analysis}
	\label{Summarization}
	In recent years, there have been several efforts to defend against adversarial attacks. We focus on the pre-processing based methods which process the inputs with certain transformations to remove the adversarial noise. In this section, we demonstrate the existing defending pre-processing based methods against adversarial attacks within the \framework in which those defenses can be concluded as perform denoising on the adversarial examples. In following Sec.~\ref{sub:spatial}, \ref{sub:frequency}, \ref{sub:latent}, the defense strategies summarized from 3 aspects (\textit{i.e.} spatial domain, frequency domain, and latent space separately). 
	
	However, most defensive methods usually focus on how to improve the defense results and rarely pay attention to the characteristic of perturbations. Unlike defensive methods that only have adversarial samples, here we make a strong hypothesis that given an adversarial input $x^{*}$, we can obtain its corresponding benign input and full perturbation $\eta$. 
	\begin{align}
	x^{*} = x + \eta
	\label{eq:perturbation}
	\end{align}

    Since all the knowledge of perturbation has been obtained. We can directly adopt a series of processing on the perturbations without sacrifice accuracy on benign images. In the section, abide by these denoise strategies we gradually remove the perturbation $\eta$ from the spatial domain, frequency domain, and latent space separately. This helps us understand the effects of attacks, and then further evaluate these defense mechanisms. As we will show, these results will shed light on how to design more robust defensive strategies. 
    
% 	Adversarial attack are often categorized as whitebox or blackbox. 
	Since the white-box attack has knowledge of the model details such as the architecture, parameters, training procedure, usually pose a more severe challenge to defense methods. Here we adopt white-box attack methods to evaluate the robustness of each defense mechanism in the worst-case scenarios.
	To ensure the generalization of the analysis results, we consider various types of attack techniques: one-step method (\textit{i.e.} FGSM~\cite{FGSM}) and iterative method (\textit{i.e.} PGD~\cite{PGD}), covering both small and large radius $\epsilon \in \{1,2,4,8\}$, measured by $l_{\infty}$ norm. Many defense techniques primarily work on smaller images, such as MNIST~\cite{MNIST} and CIFAR-10~\cite{CIFAR-10}. To reveal whether these defense strategies are still effective for complex images, here we randomly sampled 10K images from the ILSVRC2012 validation set~\cite{ImageNet}, then generate corresponding adversarial examples. Both attack and robustness evaluation is performed on the widely used classifier: ResNet-152~\cite{ResNet-152}.
	
	\begin{figure}[t]
		\begin{center}
			\includegraphics[width=1\linewidth]{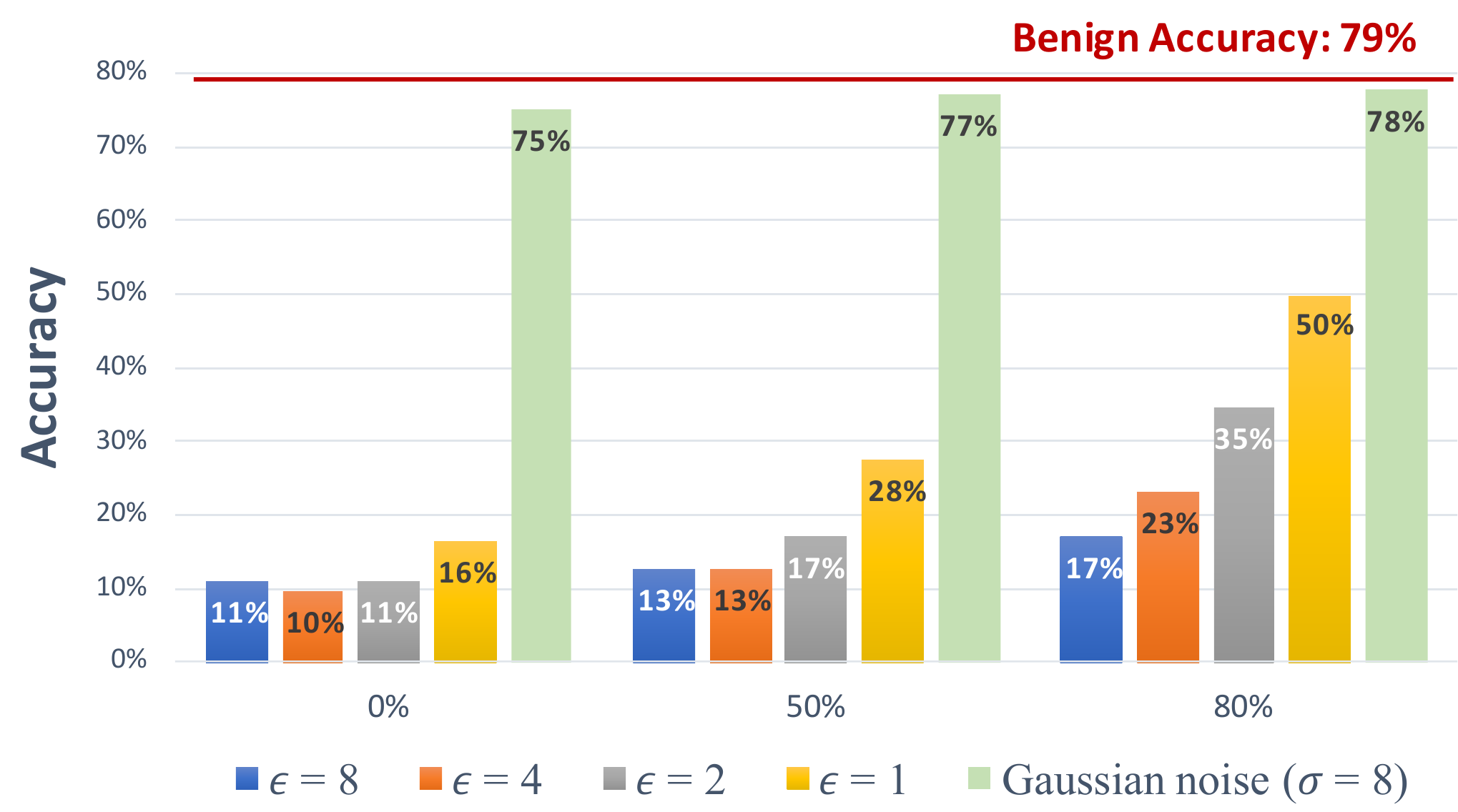}
		\end{center}
		\caption{Robustness accuracy of eliminating certain ratios of perturbations in the spatial domain.}
		\label{fig:Spatial1}
	\end{figure}
%	\vspace{1cm}
	
	\subsection{Denoising in the spatial domain}
	\subsubsection{Summarization}
	\label{sub:spatial}
	Defenses that process the inputs by spatial transformations are extensively investigated methods against adversarial attacks. Literally, the insight given by those methods is removing the adversarial noise on the image with a denoising function in the spatial domain. Such a function is denoted as $D: x^* \xrightarrow{} \hat{x}$, where $\hat{x}$ represents the denoised image. To the best of our knowledge, $D$ can be any kind of transformation driven by image quilting~\cite{image-quilting}, total variance minimization~\cite{RUDIN1992259}, pixel manipulation, deep neural network(DNN) or Generative Adversarial Network(GAN)~\cite{goodfellow2014generative}.
	
	Guo \etal~\cite{qualiting} transform the images via non-differentiable image pre-processing or randomized transformation including image quilting~\cite{image-quilting}, total variance minimization~\cite{RUDIN1992259}, and quantization. A similar idea of denoising the input in the spatial domain was later explored in~\cite{D3}, where the author proposed a framework for mitigating the effectiveness of adversarial examples by dividing the input into patches with overlapping, denoising each patch, and then reconstructing them. Prakash \etal~\cite{pixel-deflection} took advantage of localization of the perturbed pixels in their defense, proposing a non-differentiable pre-processing of inputs from another perspective. Some pixels of the image are randomly replaced with near-by pixels. 
	
	Meanwhile, there are series of adversarial defenses performed in the spatial domain which are mainly based on denoisers powered by DNNs or GANs. Gu and Rigazio~\cite{DCN} first proposed the use of denoising auto-encoders for defending. Typically, an autoencoder for denoising is trained to encode adversarial examples to clean images in order to remove the perturbations resulted from adversarial attacks. Later Meng and Chen \cite{magnet} proposed a two-step defense pipeline, which detects the adversarial inputs and then reconstructed the adversarial examples by adding Gaussian noise or encoding them with an autoencoder. \cite{DAPAS, Comdefend, APE-GAN, Defense-GAN} leveraged generative models to purify adversarial images by dragging them back towards the distribution of clean images, which are also done in the spatial domain.
	\begin{figure}[t]
		\begin{center}
			\includegraphics[width=1\linewidth]{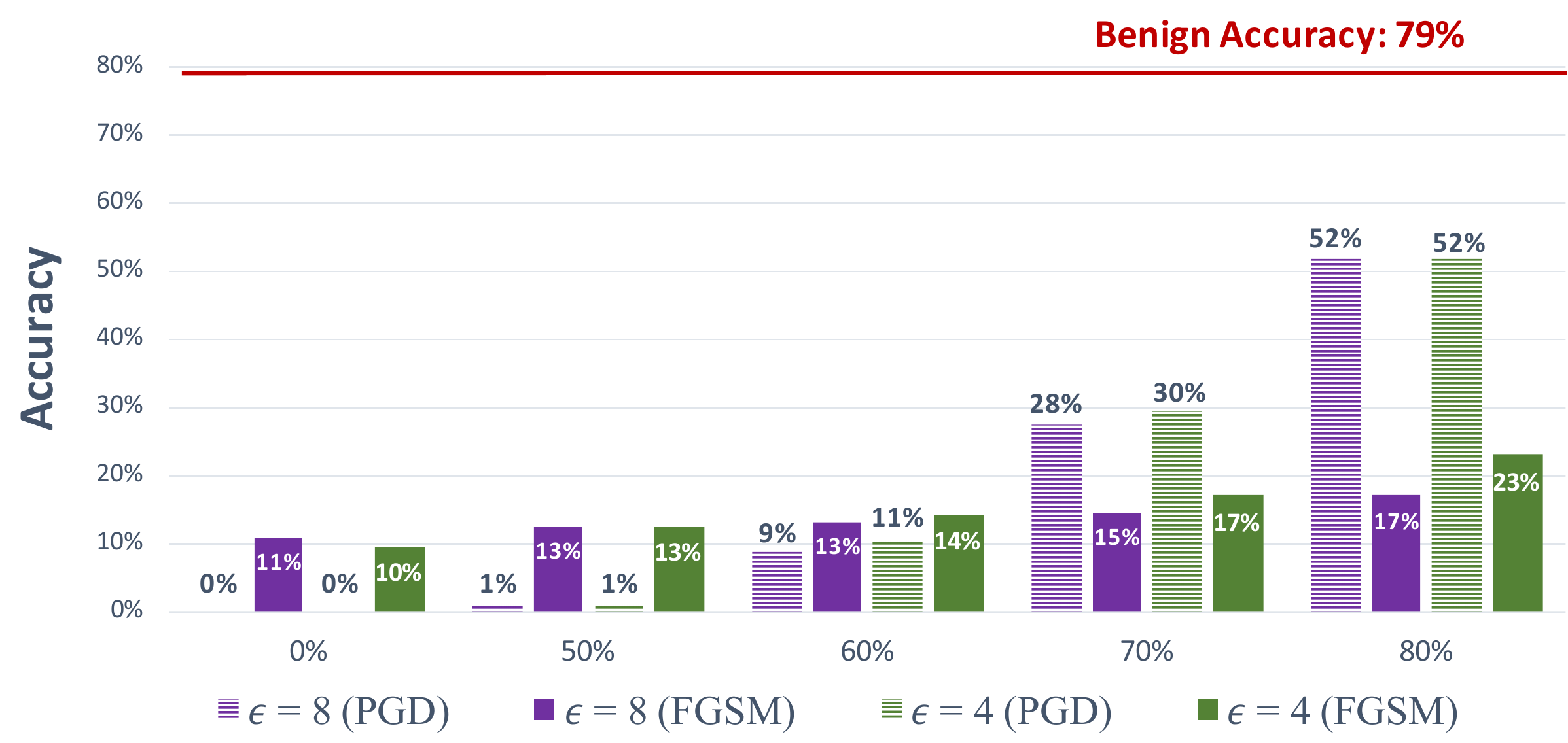}
		\end{center}
		\caption{The comparison between one-step and iterative attack methods after eliminating proportions of perturbation.}
		\label{fig:Spatial2}
	\end{figure}
	
	\subsubsection{Robustness evaluation and analysis}
	\label{sub:spatial_analysis}
	Following the idea of denoising strategy, we randomly eliminate perturbations $\eta$ according to a series of proportions, helping us quantify the impact of perturbation on model accuracy in the spatial domain. Here we set the ratios from 30\% to 80\%. Further, increase the ratio to 100\%, we will finally get the corresponding benign examples. However, for most defense methods, it is very difficult to remove all perturbations precisely without sacrifice benign accuracy, especially for complex images like ImageNet. 
	%However this has not be achieved by previous works, and it is without modifying benign inputs.  
	
	 As shown in Figure~\ref{fig:Spatial1}, we remove proportions of perturbations generated by FGSM in the spatial domain. Even if half of the perturbations are eliminated, there is almost no robustness improvement at all (for $\epsilon = 8,4$). The rest still has the same attack ability as the full perturbations. As we further increase the ratio to 80\%, most of the perturbations are eliminated. Nevertheless, the result is not as we expected. For the large radius $\epsilon = 8$, the accuracy only slightly increased from 11\% to 17\%, the remaining perturbations still have strong attack capabilities. Although the robustness accuracy is improved for a small radius ($\epsilon = 1$), there is still a big gap (nearly 30\%) from the benign accuracy.
	
	The one-step (FGSM) is less effective than iterative (PGD) attacks since iterative attacks have a higher success rate under white-box conditions. To compare the characters of one-step and iterative methods, we draw the denoising robustness under each attack in Figure~\ref{fig:Spatial2}. However, noise reduction in the spatial domain can hardly degenerate the FGSM attack effect. This is mainly because FGSM tends to generate larger distortions. When the same proportion remains, the larger distortion performed a stronger attack effect. We also try to inject Gaussian noise to the benign inputs. No matter how the ratio changes, there will be no obvious effect on model performance.
	
	According to~\cite{qualiting}, we also try to randomly remove perturbations patch-wise. However, it can not have any effect on improving defensive results. The above experiments illustrate a fact that the perturbations are redundant, only 50\% of it can achieve an attack effect consistent as all. At the same time, adversarial perturbations are far more powerful than we expected, even if the defense methods make every effort to remove most of the perturbations, the remaining parts (20\% of it) still have strong attack effects, which degrade the accuracy significantly (from 79\% to 17\%). 
	\begin{figure}[t]
		\begin{center}
			\includegraphics[width=0.9\linewidth]{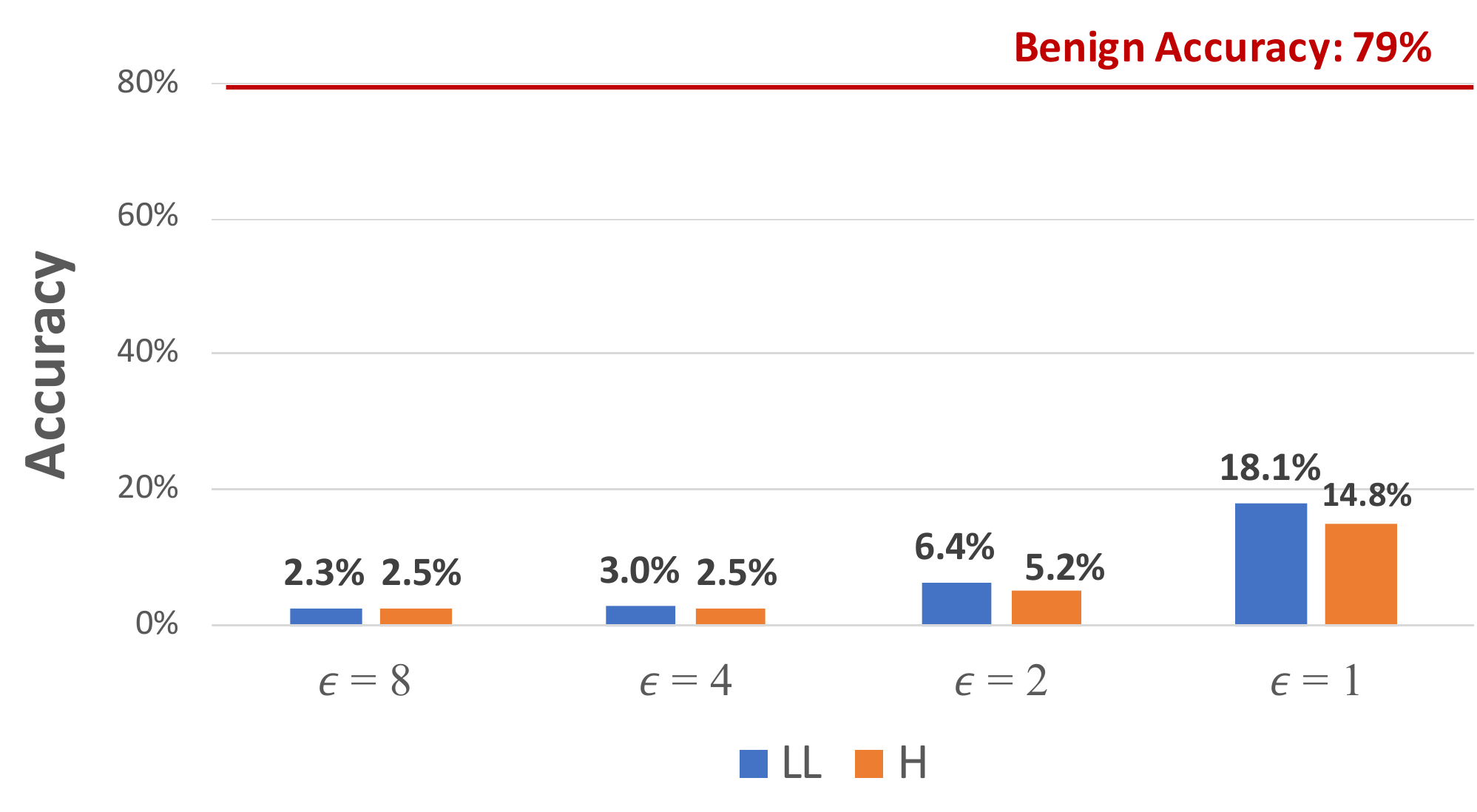}
		\end{center}
		\caption{Robustness accuracy after the perturbations of  low (LL) or high (H) frequency bands are eliminated.}
		\label{fig:Frequency1}
	\end{figure}
	
	\subsection{Denoising in the frequency domain}
	\subsubsection{Summarization}
	\label{sub:frequency}
	Another family of adversarial defenses by the pre-processing process is based on the transformations performed in the frequency domain. Most of them are under the assumption high-frequency channels contribute more to adversarial perturbations. Based on this assumption, the following defenses have borrowed several existing techniques and used them towards adversarial robustness. Frequency domain filters are introduced in \cite{Medianfilter,Low-pass} to mitigate adversarial effects. The similar idea goes for the use of JPEG compression\cite{image-quilting,Feature-Distillation,JPEG1-study,JPEG2-keep,JPEG3-Shield} as adversarial defenses, which takes the advantage of Discrete Cosine Transform (DCT) in JPEG compression to denoise the high-frequency components. More complex implementations of such a denoising transformation in the frequency domain are always possible (e.g. Saak transform\cite{Saak-Transform}), but the insights behind are common.
	
	Mustafa \etal~\cite{SR} presents a method employing super-resolution to defense classifiers against adversarial attacks. Generally, the idea is using the super-resolution as a mapping function to project the adversarial examples which lie off the manifold of low-resolution natural images in the distribution of natural images with high resolution. And visualization on magnitude spectrum shown that, in the frequency domain, high-frequency adversarial noise is removed and high-frequency components with semantic meaning are recovered (e.g. texture) by the super-resolution reconstruction.
	\begin{table}[]
		\begin{center}
			\resizebox{0.4\textwidth}{!}{
				\begin{tabular}{cccc|cccc}
					\hline
					\multicolumn{4}{c|}{Remove Specific Bands} & \multicolumn{4}{c}{PGD Attack} \\
					LL& LH&  HL& HH & $\epsilon = 8$& $\epsilon = 4$ & $\epsilon = 2$ & $\epsilon = 1$ \\ \hline
					\ding{55}&- & - & - & 2\%& 3\% & 6\%& 18\% \\
					-&\ding{55}& - & - & 0\%& 0\% & 1\%& 2\% \\
					-& - &\ding{55}& - & 0\%& 0\% & 1\%& 2\% \\
					-& - & - &\ding{55}& 0\%& 0\% & 1\%& 1\% \\
					\ding{55}& \ding{55}&\ding{55}&\ - & \textbf{70}\%& \textbf{70}\% & \textbf{70}\%& \textbf{72}\% \\      
					\hline
			\end{tabular}}
		\end{center}
		\caption{Robustness accuracy\protect\footnotemark[1] after specific frequency bands of perturbations are removed.}
% 		\footnote
		\label{tab:Frequency-combined}
	\end{table}
	\footnotetext[1]{Robustness accuracy is evaluated under PGD attack. Similar results are also observed on FGSM, please refer to the supplementary materials.}
	
%	\subsection{Perturbations in the frequency domain and robustness analysis}
	\subsubsection{Robustness evaluation and analysis}
	\label{sub:frequency_analysis}
	Most of the previous methods are based on the assumption that perturbations are generally high-frequency details. And try to improve the robustness by suppressing the high-frequency component of adversarial noise. In order to verify the contribution of high and low-frequency components to the adversarial perturbations, here we directly split perturbations into various frequencies. Then remove the corresponding frequency bands of adversarial noise and viewing the changes in the robustness accuracy.
	
	 The discrete wavelet transform (DWT) is applied, since wavelet functions are localized, a wavelet basis is often better than the Fourier basis at representing non-smooth signals. Perturbations then split into 4 sub-bands: LL (average), HL (vertical), LH (horizontal) and HH(diagonal) information. High-frequency information is mainly is distributed in H (composed of HL, LH and HH sub-bands), while low frequency is distributed in the LL band. By performing the inverse DWT (IDWT), we can obtain an output that is exactly the same as the input.

	The robustness accuracy is illustrated in Figure~\ref{fig:Frequency1}, after we remove the high (H) and low (LL) frequency bands of the perturbation respectively. It can be easily found that only discarding a certain frequency band is not enough to improve robustness. As radius $\epsilon$ increased, the adversarial effect has been strengthened. The remaining frequencies still have strong attack ability. 
	
	Instead of deleting only a certain frequency band, we further try to eliminate the combined frequency band (see Table~\ref{tab:Frequency-combined}). It is not difficult to find removing only part of the frequency band is difficult to obtain a better defense effect. When both high and low frequencies are processed, the robustness can be improved dramatically (up to 72\%)! As many works purify their adversarial examples only by filtering out high-frequency components, we suggest that to achieve an approving defense efficiency, the defense should be carried out simultaneously on multiple frequency bands.  
	% XX sup中补充 FGSM 的 results 
	\begin{figure}[t]
		\begin{center}
			\includegraphics[width=0.9\linewidth]{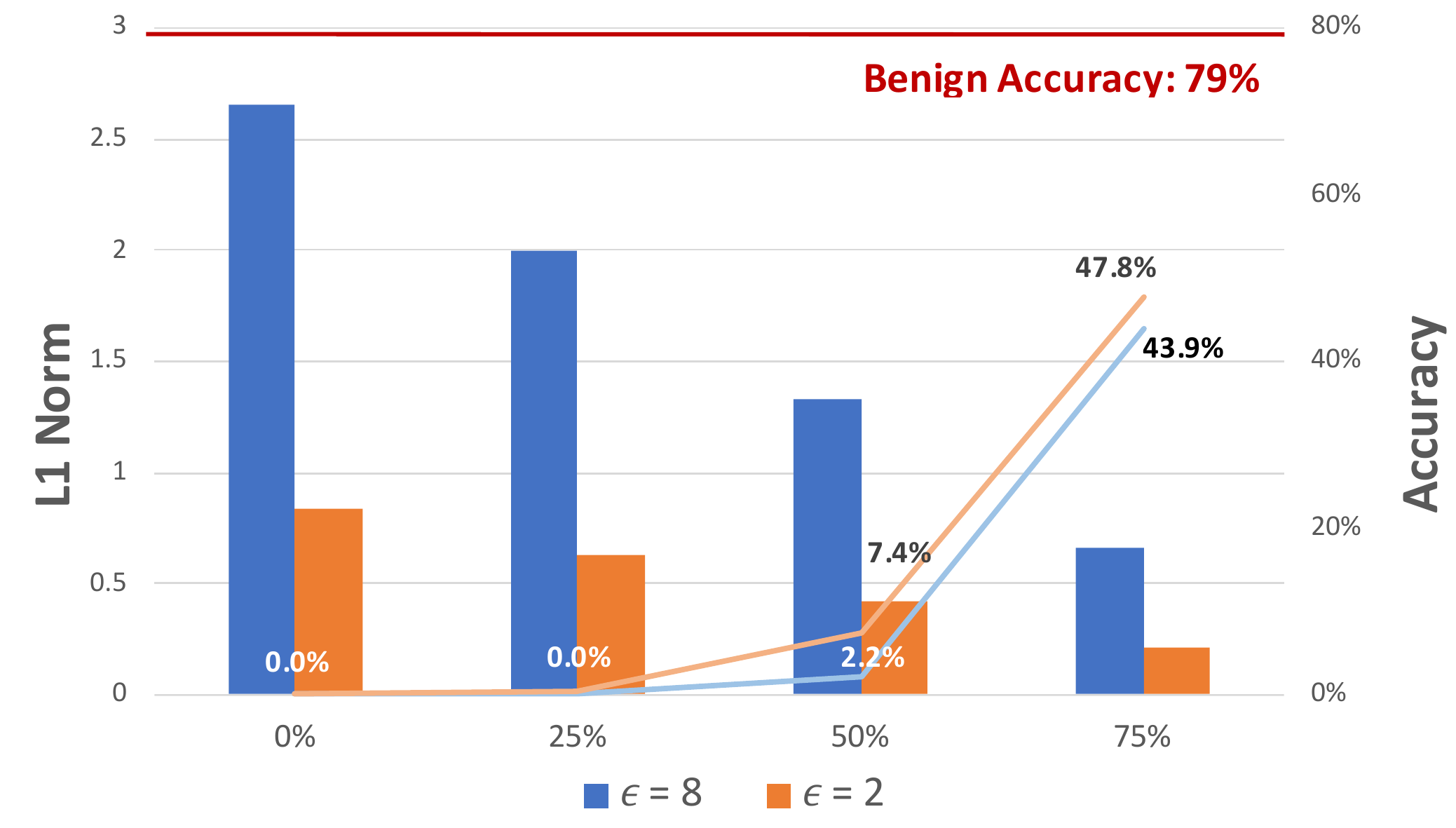}
		\end{center}
		\vspace{-0.3cm}
		\caption{Eliminate the $\eta_{\,latent}$ in channel-wise.}
		\label{fig:Latent-channel}
		\vspace{-0.3cm}
	\end{figure}

	\subsection{Denoising in the latent space}
	\subsubsection{Summarization}
	\label{sub:latent}
	Another segment of approaches\cite{Feature-Distillation,HGD,NRP,STL} to increase the robustness of the model against adversarial attacks, which are also powered by DNN, are used to be treated in a partial perspective. Within the \framework, they should be distinguished from spatial domain methods\cite{DCN,magnet,DAPAS,Comdefend,APE-GAN,Defense-GAN} mentioned in Sec. \ref{sub:spatial}, as the key steps of them are done in the latent space (e.g. heatmaps or intermediate representations).
	
	Such an approach was first introduced by Liao \etal~\cite{HGD}, where they proposed a High-level Representation Guided (HGD) denoiser, given the observation that perturbations seem small in images but they are amplified in higher representations - the latent space.
	Sun \etal~\cite{STL} designed a sparse transformation layer (STL) to map the input images to a low dimensional quasi-natural space. Naseer \etal~\cite{NRP} proposed an approach to denoise and reconstruct images in the perceptual feature space by a Purifier Network. And \cite{Feature-Regeneration} leveraged resilient feature regeneration defense which can outperform existing state-of-the-art defense strategies according to the paper. The main idea is to deal with the vulnerability of individual convolutional filters in DNNs, which reveals the significant impact of adversarial noise in the latent space.
	
		\begin{figure}[t]
		\begin{center}
			\includegraphics[width=0.9\linewidth]{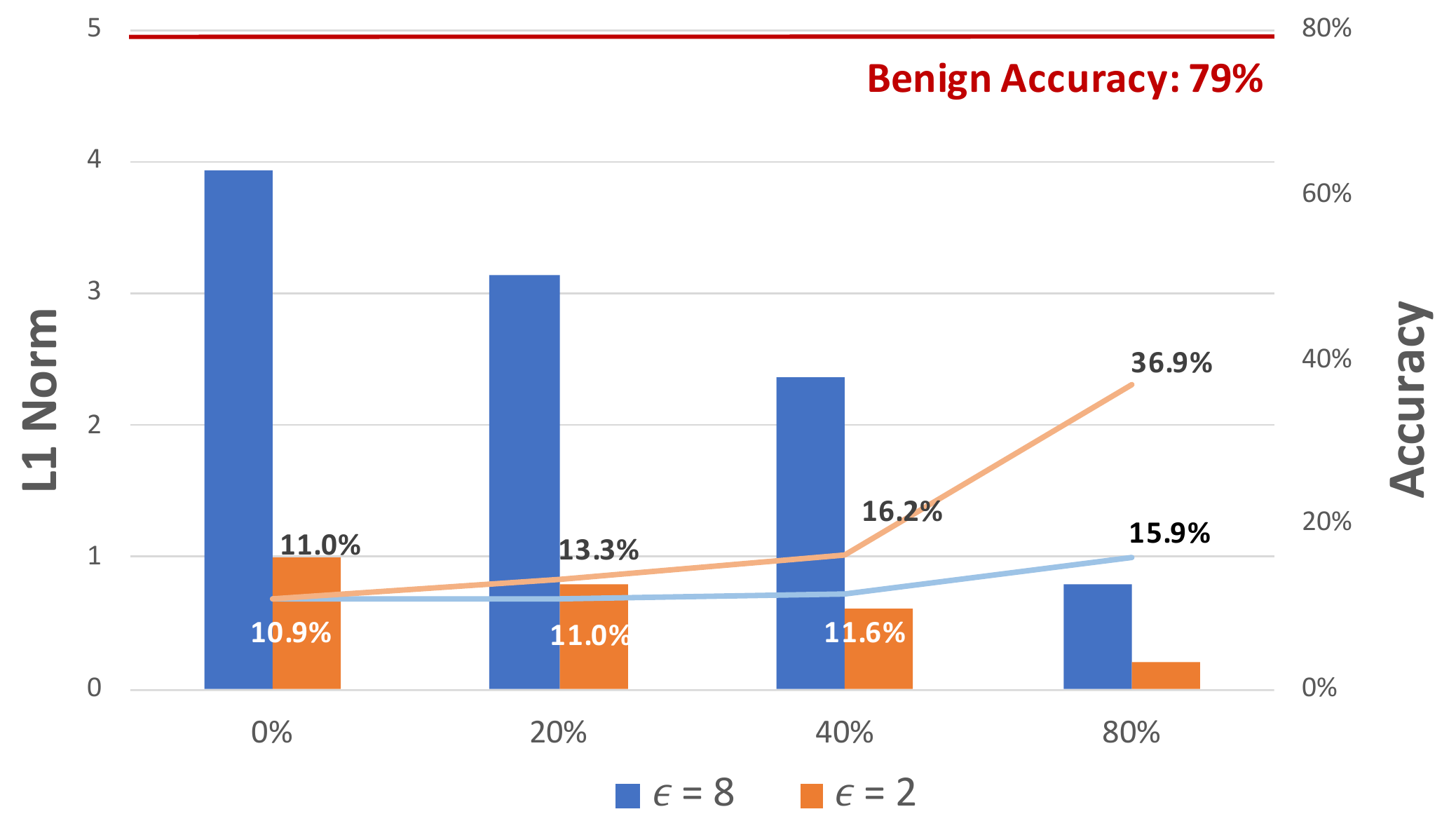}
		\end{center}
		\vspace{-0.3cm}
		\caption{Eliminate the $\eta_{\,latent}$ in spatial-wise.}
		\vspace{-0.3cm}
		\label{fig:Latent-spatial}
	\end{figure}
	
	\begin{figure*}[t]
		\begin{center}
			\includegraphics[width=0.9\linewidth]{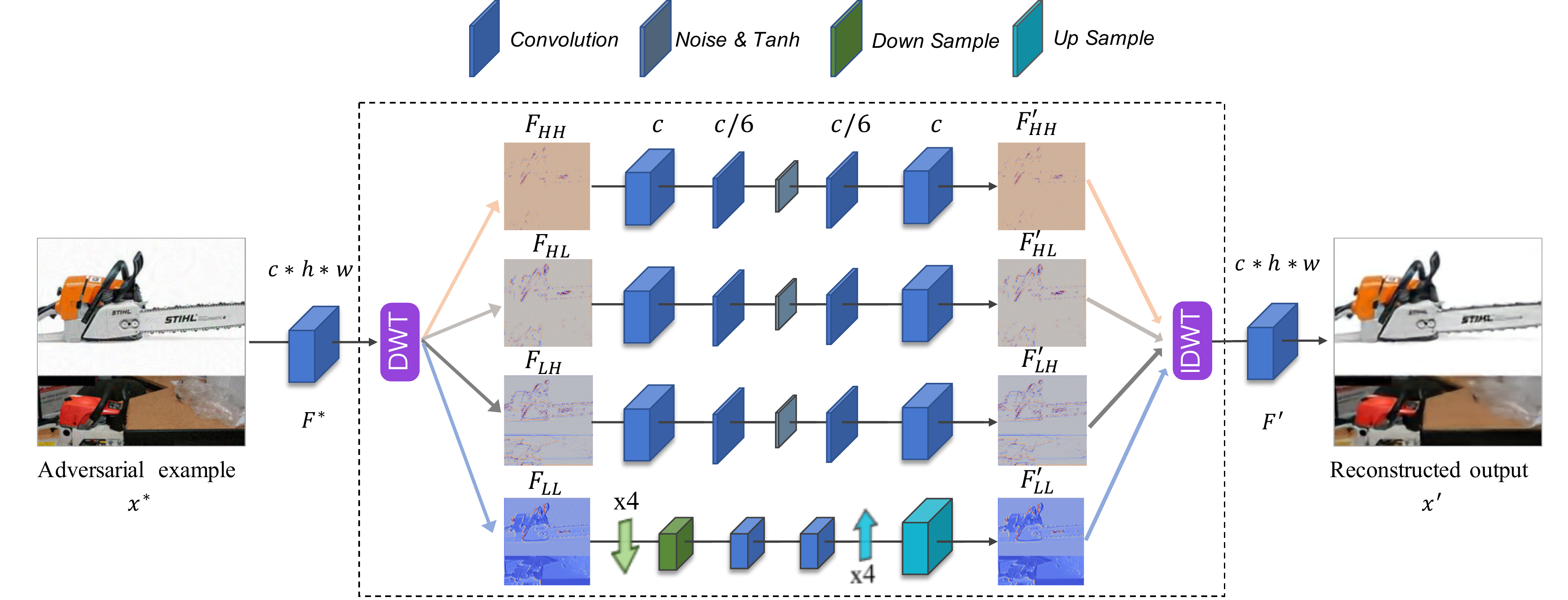}
		\end{center}
% 		\vspace{-1em}
		\caption{The architecture of ACM. Adaptive compression is performed according to the characteristics of high and low-frequency bands. For the high-frequency features $(F_{HH}$, $F_{HL}$, and $F_{LH})$, a channel-wise compression is adopted, and the spatial compression is carried out for the low-frequency features $(F_{LL})$.}
		\label{fig:Model}
% 		\vspace{-0.2cm}
	\end{figure*}
	
	\begin{figure}[h]
		\begin{center}
			\includegraphics[width=0.9\linewidth]{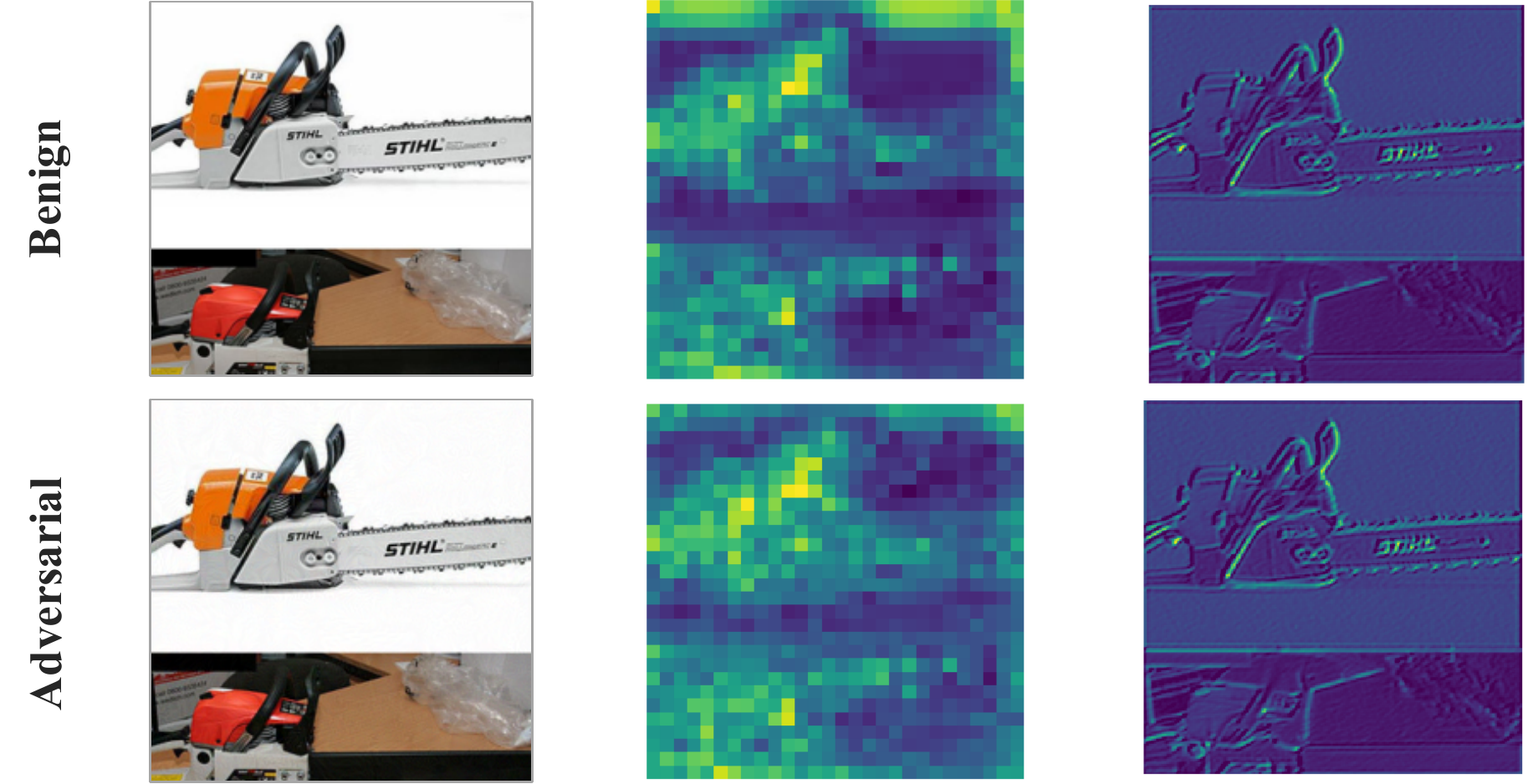}
		\end{center}
		\caption{Feature maps corresponding to a benign image (top) and adversarial perturbed counterpart (bottom). Middle: Extracted by $res_{2}$ block of ResNet-152. Right: Extracted by \framework.}
		\label{fig:Feature-maps}
	\end{figure}
	
%	\subsection{Perturbations in the latent space}
	\subsubsection{Robustness evaluation and analysis}
	\label{sub:latent_analysis}
	As summarized in Sec.~\ref{sub:latent}, these defense methods make efforts to denoising perturbations in the latent space. However, most of them usually have a strong correlation with specific classification models and attack methods. From a more general perspective, we evaluate this strategy under the \framework, ensuring agnostic to attack methods and models. we first construct an AutoEncoder-based model following Comdefend~\cite{Comdefend}, then extract features by the encoder module $E(\cdot)$. The behavior of perturbations in latent space can be represented as 
	\begin{align}
	\eta_{\,latent} = E(x^{*}) - E(x)
	\end{align}
	Here the size of $\eta_{\,latent}$ is $C\times H \times W$ .To evaluate the robustness of the denoising strategy in latent space. In the following content, we eliminate the adversarial impact by reducing the $||\eta_{\,latent}||_{1}$ gradually. Here we decrease $||\eta_{\,latent}||_{1}$ by randomly setting the value of $\eta_{\,latent}$ to 0 in two manners: channel-wise and spatial-wise respectively. For iterative attacks (PGD), robustness accuracy is illustrated in Figure~\ref{fig:Latent-channel}. As we increase the ratio, the $L1$ distance is significantly reduced. However, when a quarter of the $\eta_{\,latent}$ is removed, there is basically no improvement in terms of accuracy. As the ratio increased to 75\%, most adversarial information $\eta_{\,latent}$ have been eliminated. However, the robustness accuracy has only increased to about 48\%, and there is still a big gap from the benign accuracy (79\%). We also remove the $\eta_{\,latent}$ for one-step attack (FGSM) in spatial-wise. (see Figure~\ref{fig:Latent-spatial}) and get similar results. For a smaller attack radius ($\epsilon=2$), even 80\% of $\eta_{\,latent}$ the is remove, the robustness accuracy is only increased by 25.9\%.

	\subsection{Analyses and suggestions}
	\label{sec:suggestion}
	As the quantitative experiments and analyses above, denoising strategies from 3 aspects both have their own limitations. It is difficult to obtain a satisfactory robustness result, even most of the adversarial perturbations are eliminated in the spatial domain. Moreover, for most defense methods, eliminating perturbation without degrading benign input is still arduous work. For the defensive strategy in the frequency domain, the characteristics of each frequency component should be considered. As the low-frequency bands mainly contain the general outline and contour, which is the approximate information of the image. High frequency mainly contains details and texture information as well as noise. Compared to defending on a signal frequency band, we recommend performing defense on both high and low-frequency bands according to their characteristics.

	As proposed in~\cite{Feature-Denoising}, adversarial attacks lead to very substantial “noise” in the feature maps which extract by classification-related models (see Figure~\ref{fig:Feature-maps} the second column). Therefore, under these circumstances, denoising can be a feasible strategy. However, the \framework mainly focuses on the reconstruction of image semantic information and is not sensitive to noise (see Figure~\ref{fig:Feature-maps} the third column). Therefore, directly denoising in feature spaces can hardly discard the adversarial perturbations. The analyses in Sec.~\ref{sub:latent_analysis} both reveal the same point. Therefore, we recommend adopting a compression strategy. First, under the constraints of compression, the semantic information of the image will be more easily reconstructed than the irregular adversarial noise. Secondly, since the compressed information is changed to some extent, even if the adversarial perturbations still exist, it can no longer complete the attack.
	\begin{table*}[]
		\begin{center}
			\resizebox{0.7\textwidth}{!}{
				\begin{tabular}{l|c|cccc|cccc|cccc}\hline
					\multirow{2}{*}{Method} & \multirow{2}{*}{Benign} & \multicolumn{4}{c|}{\multirow{2}{*}{FGSM\cite{FGSM}}} & \multicolumn{4}{c|}{\multirow{2}{*}{MI-FGSM\cite{MI-FGSM}}} & \multicolumn{4}{c}{\multirow{2}{*}{PGD\cite{PGD}}} \\
					&                         &  \multicolumn{4}{c|}{}    & \multicolumn{4}{c|}{}      &\multicolumn{4}{c}{}    \\ \hline
					No defense               & \textbf{78.9} &  16.4   & 11 & 9.6  & 10.9  & 3.3 & 1.1 & 0.6  & 0.3 & 0.5 & 0.2 & 0.1 & 0.0\\
					JPEG\cite{JPEG2-keep}    & 75.1 &  59.9   & 46.4 & 30.9  & 22.3  & 57.8 & 44.8  & 24.2  & 6.3 & 63.7 & 56.0 & 45.5 & 27.1\\
					FD\cite{Feature-Distillation}  & 75.1&  57.4   & 43.3 & 27.7  & 18.7  & 54.4 & 37.3  & 17.2  & 3.9 & 61.8 & 52.2 & 38.9 & 19.6\\
					WD+SR\cite{SR}   & 74.2 &  44.9   & 31.8 & 24.4  & 22.4  & 38.1 & 20.9  & 7.5  & 4.1 & 43.8 & 27.6 & 14.9 & 7.1\\
					NRP\cite{NRP}  & 74.1 &  50.5   & 36.8 & 27.7  & \textbf{55.1}  & 46.2 & 29.7  & 10.8  & 6.1 & 51.6 & 37.7 & 23.8 & 21.2\\
					PD\cite{pixel-deflection}  & 74.9 &  49.4   & 31.5 & 18.0  & 12.9  & 45.7 & 27.8  & 11.3  & 2.1 & 51.1 & 35.3 & 19.2 & 7.1\\
					AT\cite{PGD}	& 45.4 &  45.3   & 45.1 & 44.7  & 43.8  & 46.8 & 46.7  & 46.7  & \textbf{45.9} & 45.3 & 45.0 & 45.0 & 44.8\\
					Ours      & 70 &  \textbf{63.4}   & \textbf{58.2} & \textbf{48.9}  & 37.3  & \textbf{64} & \textbf{58.2} & \textbf{48.6} & 35.4 & \textbf{65.7} & \textbf{62.5}  & \textbf{59}  & \textbf{53.7}\\ \hline
				% 	Ours*                         & 69.5 &  \textbf{63.5}   & \textbf{59} & \textbf{50.6}  & 37.8  & \textbf{64.7} & \textbf{59.4} & \textbf{51.1} & 37.9 & \textbf{64.8} & \textbf{61.3}  & \textbf{59.5}  & \textbf{54.8}\\  \hline
				% 	HGD                         & 0 &  01   & 01 & 01  & 01  & 01 & 01  & 01  & 01 & 01 & 01 & 01 & 01\\
			\end{tabular}}
		\end{center}
		\caption{The robustness accuracy (\%) compared with other defensive methods ($L_{\infty}=$ 1 / 2 / 4 / 8).}
		\label{tab:robustness accuracy}
	\end{table*}
	%-------------------------------------------------------------------------
	\section {Adaptive Compression in the Feature Domain}
	As summarized in Sec.~\ref{sec:suggestion}, for the \textit{F-IF} defense framework, we recommend adopt adaptive compression for various frequency bands. Following this idea, we construct an Adaptive Compression and reconstruction model (ACM). As shown in Figure~\ref{fig:Model}, give an adversarial example $x_{adv}$, multi convolution layers are used to extract features 
	\begin{align}
	F^{*} = H(x^{*})
	\end{align}
	where $H(\cdot)$ denotes multi convolution operations. Here we set channel $c=32$, $h$ and $w$ are the height and width of the feature maps. Then the discrete wavelet transform (DWT) is introduced to split them into various frequency bands. 
	\begin{align}
	F_{HH}, F_{HL}, F_{LH}, F_{LL} = DWT(F^{*})
	\end{align}
	where $F_{HH}$, $F_{HL}$ and $F_{LH}$ denote the high-frequency features $F_{H}$, which mainly contains information such as texture and details. $F_{LL}$ represents the low-frequency features. For the high-frequency features, we apply channel-wise compression and then reconstruct the main of it. 
	\begin{align}
	F^{'}_{H} = H_{Rec}(Tanh(H_{Com}(F_{H})))
	\end{align}
	where $H_{Rec}$ and $H_{Com}$ denote the compression and reconstruction respectively. During the compression process, the channel is compressed to $\frac{1}{6}$ of its original size. 
	
\begin{figure}[t]
		\begin{center}
			\includegraphics[width=0.95\linewidth]{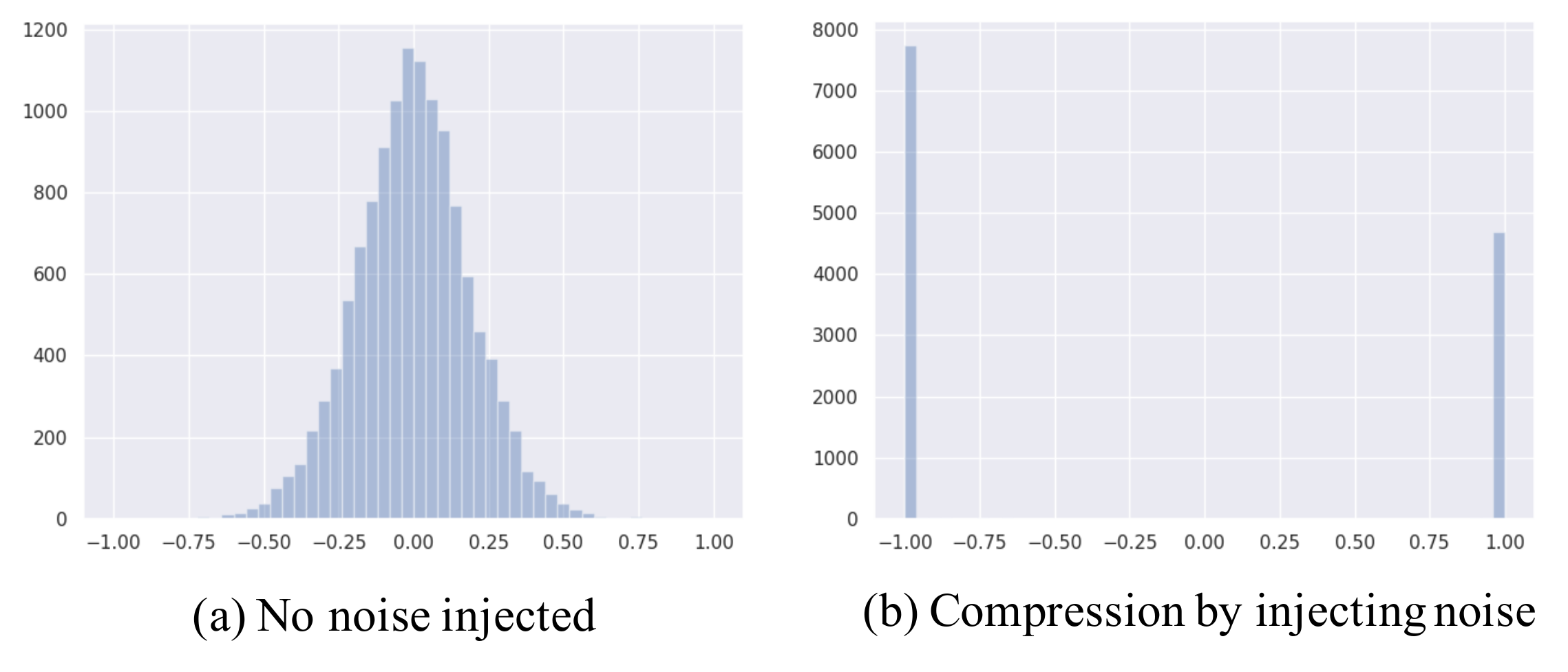}
		\end{center}
		\caption{Further compress the high-frequency bands with injecting gaussian noise before Tanh activation.}
		\label{fig:Tanh}
		\vspace{-0.3cm}
	\end{figure}

	To further compress the high-frequency bands, we also try to inject gaussian noise before Tanh activation in the compression stage. It can be easily found that when no noise is added (Figure~\ref{fig:Tanh}a), the features are widely distributed in $(-1,1)$. While the adversarial perturbations are more likely to be rebuilt out. After adding noise, the features are maximally compressed, and most values are compressed around -1 or 1.
	
	Since low-frequency features $F_{LL}$ usually contain most of the semantic information, compress in channel-wise directly will inevitably lead to information loss. Therefore, for this frequency band, we alleviate the perturbation effects via a down-sample operation. After the multi-layer convolution,  upsampling is performed to ensure the size consistent with the input. 
	\begin{align}
	F^{'}_{LL} = H_{Up}(H_{Rec}(H_{Down}(F_{LL}))))
	\end{align}
	where $H_{Down}$ and $H_{Up}$ denote the down-sample and up-sample operation. Multi convolution layers $H_{Rec}$ are adopted to reconstruct the down-sample features. After we get the processed features, the inverse discrete wavelet transform (IDWT) is deployed to obtain the reconstruct feature maps $F^{'}$. 
	\begin{align}
	F^{'} = IDWT(F^{'}_{HH}, F^{'}_{HL}, F^{'}_{LH}, F^{'}_{LL})
	\end{align}
	Finally, through a layer of convolution, we got the output which more close to the benign images. The adversarial perturbations are eliminated by the adaptive compression. 
	
	\vspace{-0.1cm}
	\section {Experiment Results}
% 	\subsection{Implementation details}
	Our model is optimized by ADAM~\cite{adam} with $L1$ Loss by setting $\beta_{1}=0.9$, $\beta_{2}=0.999$ and $e=10^{-8}$. The batch size is set to 64, and the learning rate is initialized to $2 \times 10^{-4}$ and halved every $4 \times 10^{5}$ Iterations. Here, training data is consists of 400 images from the ILSVRC training set.

%\linespread{1}
	\subsection{Robustness accuracy}
	We compared our method with 6 state-of-the-art defense mechanisms : JPEG~\cite{JPEG2-keep}, Feature Distillation (FD)~\cite{Feature-Distillation}, WD+SR~\cite{SR}, NRP~\cite{NRP}, Pixel Deflection (PD)~\cite{pixel-deflection}, Adversarial Training with PGD (AT)~\cite{PGD}. Attack techniques adopted including Fast Gradient Sign Method (FGSM)~\cite{FGSM}, PGD~\cite{PGD}, Momentum Iterative FGSM ~\cite{MI-FGSM}. Both attack and evaluation are performed on the 1000 images from ILSVRC validation set. Robustness accuracy is summarized in Table~\ref{tab:robustness accuracy}. As can be seen, both MI-FGSM and PGD attacks were successful at reducing the classifier accuracy by up to 78\%. Benefit from the adaptive compression strategy, our ACM is more robust than other methods. For the small radius, adversarial noises are imperceptible to human eyes, our method outperform existing defense strategies. In particular, it improves the robustness accuracy from 0\% to 65.7\% under the strong PGD attack.

	For most of the defense methods, thousands of benign images and their perturbed counterparts are necessary to learn how to eliminate specific adversarial perturbations. Compared with the adversarial training and other methods, our ACM has much lower training overhead, only hundreds of benign images are enough. Moreover, our method is agnostic to attacks and models, easily combined with other defense to improve robustness without modifying the classifier. 
% 	Therefore, our adaptive compression strategy effective defense against adversarial example.

	\begin{figure}[t]
		\begin{center}
			\includegraphics[width=0.9\linewidth]{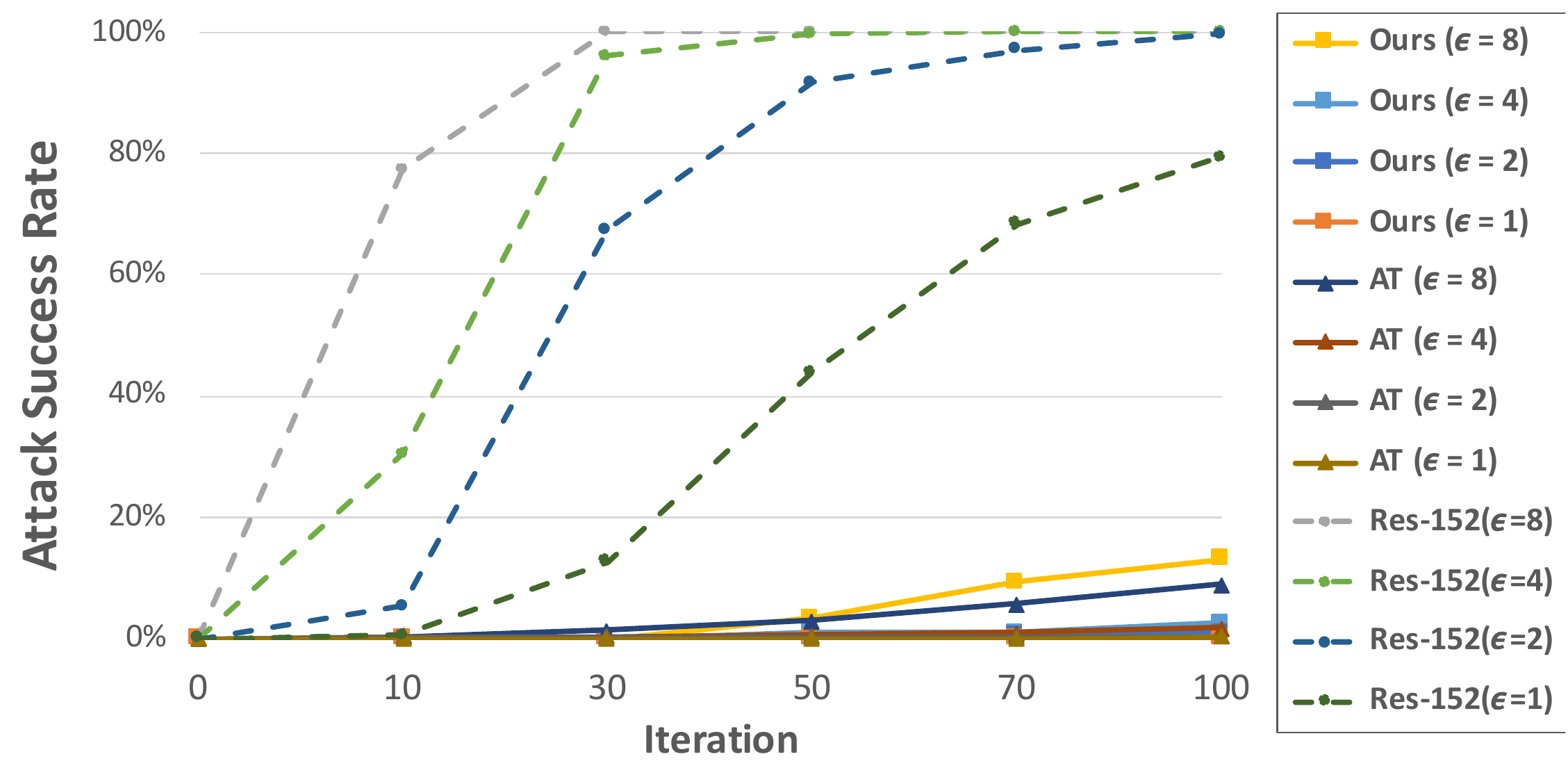}
		\end{center}
		\caption{The success rate of adaptive attack.}
		\label{fig:Adaptive-attack}
	\end{figure}
	
	\begin{figure}[t]
		\begin{center}
			\includegraphics[width=0.95\linewidth]{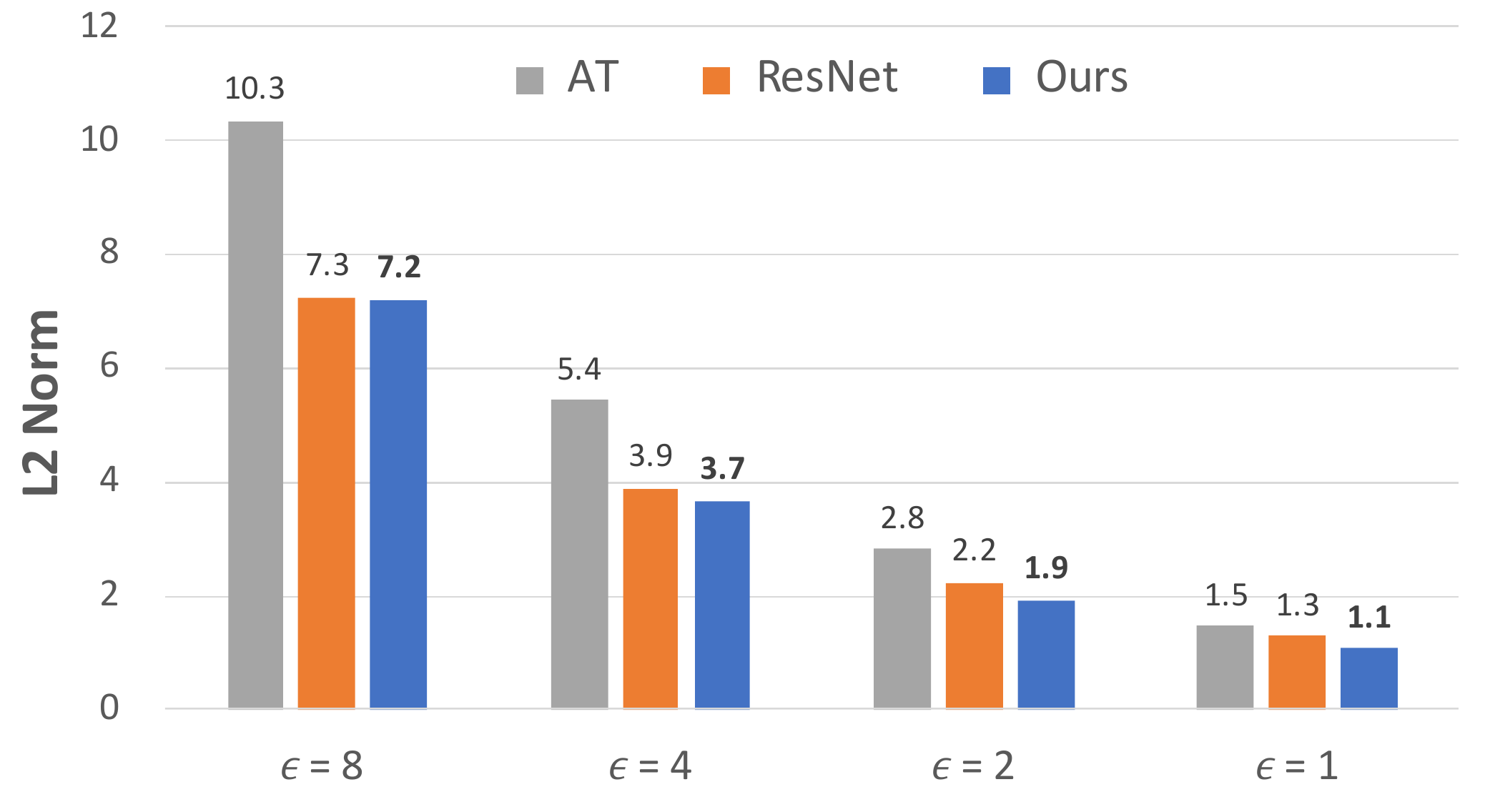}
		\end{center}
		\caption{The $L2$ norm of perturbations (generated under $L_{\infty} \in \{1,8\}$  distance metrics).}
		\label{fig:Adaptive-distance}
	\end{figure}
	
	\subsection{Robustness against adaptive attack}
	To analyze the model robustness against adaptive attacks, we allow an adversary to have full access to our model. Here, we perform a strong attacker (PGD) with up to 100 attack iterations, where the targeted label is selected uniformly at random. We evaluate with $\epsilon \in [1,8]$, cover a wide range of adversaries.
	
	Without any defense, for large radius $\epsilon$ the attack success rate on ResNet-152 achieve almost 100\% after 30 iterations (dotted line in Figure~\ref{fig:Adaptive-attack}). Compared with the adversarial training model (AT), under the 50-iteration attack, the attack success rate of AT and ours are 3\% and 3.3\% respectively. Results on adaptive attack demonstrate the robustness of our adaptive compression strategy. 
% 	More details about the robustness accuracy under adaptive attack can be seen in supplementary materials. 
	
	To further understand our adaptive compression strategy, we observe the generated perturbations under 100-Iteration in $L2$ metric (see Figure~\ref{fig:Adaptive-distance}). For the adversarial training model (AT), without the adaptive compression strategy, the attacker tends to large perturbation to attack model, robustness accuracy is 39.7\% ($\epsilon = 4$). By contrast, under the same constraints, the perturbation generated by the attacker is relatively smaller, and robustness accuracy is 58.7\% ($\epsilon = 4$). It shows that the attacker cannot achieve a better attack effect only by generating a large distortion. Our adaptive compression strategy makes it more difficult to fool the model. More details about the robustness accuracy are listed in the supplementary material.

	\subsection{Ablation study}
	To verify the effectiveness of adaptive compression on all frequency bands, we perform an ablation experiment. The robustness accuracy is presented in Table~\ref{tab:Ablation study}. As shown in the first row, the model performs the best results when we perform defend on both high (HH, HL, and LH) and low (LL) frequency bands. However, if the frequency band is partially compressed, it will lead to the degradation of its robustness. When there is no defense against the LL band, the performance drop is up to -19.1\%. The largest degradation occurs (-23.5\%) when there is no defense against both HH, HL, and LH bands. 
	
	The above results demonstrate that only defending on a certain frequency band is sub-optimal. To achieve a better defense effect, adaptive compression should be performed on all frequency bands according to its characteristics.
% 	We analyze the combination of high and low bands,
    \linespread{1.2}
	\begin{table}[]
	\begin{center}
		\resizebox{0.4\textwidth}{!}{
			\begin{tabular}{cccc|cccc}
				\hline
				\multirow{2}{*}{HH} & \multirow{2}{*}{HL} & \multirow{2}{*}{LH} & \multirow{2}{*}{LL} & \multicolumn{4}{c}{Robustness accuracy (\%)} \\
				&                   &                   &                     & $\epsilon = 1$& $\epsilon = 2$ & $\epsilon = 4$& $\epsilon = 8$\\ \hline
				\ding{51}& \ding{51}&\ding{51}&\ding{51}& \textbf{65.7}& \textbf{62.5} & \textbf{59.0}& \textbf{53.7}\\ \hline   
%w/o HH, HL, LH
				\ding{55}& \ding{51}&\ding{51}&\ding{51}& -4.7& -11.0 & -16.9& -15.1\\
				\ding{51}& \ding{55}&\ding{51}&\ding{51}& -7.9& -16.5 & -17.1& -10.7\\
				\ding{51}& \ding{51}&\ding{55}&\ding{51}& -7.3& -14.4 & -19.3& -16.2\\
%w/o LL
				\ding{51}& \ding{51}&\ding{51}&\ding{55}& -10.6& -19.1 & -17.9& -15.7\\
%w/o H				
				\ding{55}& \ding{55}&\ding{55}&\ding{51}& -10.7& -19.3 & -23.5& -23.0 \\  
				\hline    
		\end{tabular}}
	\end{center}
	\caption{Ablation study of the adaptive compression performed on multi-frequency bands.}
	\label{tab:Ablation study}
\end{table}
\linespread{1}

	\section{Conclusion}
	As machine learning models are known to be vulnerable to adversarial attack, many denoising-based defense methods have been proposed. In this study, we summarize and analyze these defenses as a symmetric transformation framework via image denoising and reconstruction (\framework). Then we categorize these denoising strategies from three aspects(denoising in the spatial domain, frequency domain, and latent space). To evaluate the defense effectiveness of these strategies, we apply them directly against perturbations. However, all the results indicate the limitations of denoising strategies. Observations and analyses shed new light on developing adversarial defenses strategies. Then we propose the adaptive compression defensive strategy, applying on both high and low-frequency bands based on its characteristics. Following this idea, we construct the adaptive compression and reconstruction model (ACM). The experiment results reveal that our defensive strategy enables the model to further eliminate the adversarial perturbation, achieving a more promising robustness effect.
	
% 	Even if most of the adversarial perturbations have been eliminated, it still hard to achieve a satisfactory robustness.
% 	\clearpage
	{\small
		\bibliographystyle{ieee_fullname}
%		\bibliography{egbib}

	}
	
\end{document}